\UseRawInputEncoding

\documentclass[runningheads]{llncs}
\usepackage{graphicx}
\usepackage{wrapfig}
\usepackage{tikz}
\usepackage{bbding}
\usepackage{pifont}
\usepackage{textcomp}
\usepackage{comment}
\usepackage{amsmath,amssymb} 
\usepackage{color}
\usepackage{subfiles}
\usepackage[pagebackref,breaklinks,colorlinks]{hyperref}
\hypersetup{citecolor=[RGB]{119,185,0}}
\usepackage{xcolor}
\usepackage{colortbl}
\usepackage{hhline}
\usepackage{multirow}
\usepackage{caption}
\usepackage{makecell}
\usepackage[width=122mm,left=12mm,paperwidth=146mm,height=193mm,top=12mm,paperheight=217mm]{geometry}
\usepackage{booktabs}
\usepackage{amssymb}
\usepackage{longtable}
\usepackage[misc]{ifsym}

\begin{document}

\pagestyle{headings}
\mainmatter

\title{Motion Gait: Gait Recognition via Motion Excitation}

\author{
    Yunpeng Zhang\textsuperscript{\rm 1},
    Zhengyou Wang\textsuperscript{\rm 1,2 \Letter},
    Shanna Zhuang\textsuperscript{\rm 1,2},
    Hui Wang\textsuperscript{\rm 2}
}
\institute{
    $^1$Shijiazhuang Tiedao University \\
    $^2$Hebei Key Laboratory for Electromagnetic Environmental Effects and Information Processing \\
}

\newcommand{\fn}[1]{\footnotesize{#1}}
\newcommand{\green}[1]{\textcolor[RGB]{96,177,87}{#1}}
\newcommand{\gbf}[1]{\green{\bf{\fn{(#1)}}}}
\def \LN {\eta\xspace}
\def\ie{\textit{i.e.}}
\def\eg{\textit{e.g.}}
\def\etc{etc}
\def\etal{\textit{et al.}}

\newcommand{\wwh}[1]{\textcolor{blue}{[wwh: #1]}}

\authorrunning{Y. Zhang et al.}

\maketitle

\begin{abstract}
	Gait recognition, which can realize long-distance and contactless identification, is an important biometric technology. Recent gait recognition methods focus on learning the pattern of human movement or appearance during walking, and construct the corresponding spatio-temporal representations. However, different individuals have their own laws of movement patterns, simple spatial-temporal features are difficult to describe changes in motion of human parts, especially when confounding variables such as clothing and carrying are included, thus distinguishability of features is reduced. In this paper, we propose the Motion Excitation Module (MEM) to guide spatio-temporal features to focus on human parts with large dynamic changes, MEM learns the difference information between frames and intervals, so as to obtain the representation of temporal motion changes, it is worth mentioning that MEM can adapt to frame sequences with uncertain length, and it does not add any additional parameters. Furthermore, we present the Fine Feature Extractor (FFE), which independently learns the spatio-temporal representations of human body according to different horizontal parts of individuals. Benefiting from MEM and FFE, our method innovatively combines motion change information, significantly improving the performance of the model under cross appearance conditions. On the popular dataset CASIA-B, our proposed Motion Gait is better than the existing gait recognition methods.
\end{abstract}




\section{Introduction}\label{Indro}
\par{
	Gait recognition, as a biometric technology that can be realized in long-distance and non-contact conditions, takes people's walking posture as the basis of identity discrimination, and the characteristics of pedestrian body shape and walking posture are difficult to be camouflaged or changed. Therefore, gait recognition can be widely used in intelligent transportation, video surveillance, public security and other fields. However, changes such as carrying a backpack, wearing a coat and camera viewing angle will change the appearance of gait, which brings great challenges to gait recognition.
}
\par{
	In order to meet these challenges, many methods based on deep learning provide promising solutions \cite{Chao2019GaitSetRG,Wu2017ACS,Shiraga2016GEINetVG,Takemura2018MultiviewLP,Zhang2019LearningJG,Zhang2019GaitRV}. Lin et al. \cite{Lin2020GaitRW} proposed the Local Transform Module to realize multi-scale change of gait sequence, so as to learn multi-scale time information. GaitGL \cite{Lin2021GaitRV} preserved spatial information through the Local Temporal Aggregation Module and integrated global and local features. GaitPart \cite{Fan2020GaitPartTP} learned the fine-grained characteristics of spatial parts and conducted short-term time modeling for different parts of the human body. GaitSet \cite{Chao2019GaitSetRG} regarded gait sequence as a set and divided pedestrian gait into different horizontal parts to extract spatio-temporal features.
}
\par{
	However, the above methods only consider the spatio-temporal characteristics of gait, take the complete sequence or temporal parts as a whole, and simply learn the spatio-temporal representations, which will cause the loss of information about pedestrian motion changes. Especially under the conditions of carrying backpacks and wearing coats, the temporal changes of pedestrian gait can provide information that is less affected by covariates, which is conducive to improving the performance of cross covariate gait recognition. Moreover, due to different horizontal parts have various appearances and motion patterns, many methods extract more detailed local information by segmenting gait features horizontally. However, these methods can not extract the unique feature representation of different local parts, which reduces the recognition performance.
}
\par{
	To solve the above problems, based on GaitGL \cite{Lin2021GaitRV} framework, we design the Motion Excitation Module (MEM), which extracts motion change information from sequence features and guides the model to focus on the elements with large dynamic changes. Furthermore, in order to extract more discriminative spatio-temporal representations of various parts of the human body, we propose the Fine Feature Extractor (FFE) to independently learn gait features of different horizontal parts.
}
\par{
	The main contributions of this paper are the following three aspects.
}
\par{
	1. We propose a novel Motion Excitation Module (MEM) to extract motion change information. MEM guides the spatio-temporal characteristics of pedestrians to pay attention to the changes of different local parts of pedestrians through motion change representations.
}
\par{
	2. We present a novel Fine Feature Extractor (FFE) to specifically learn local gait features to obtain more discriminative gait feature representations of different parts.
}
\par{
	3. The proposed method is evaluated on CASIA-B dataset \cite{Yu2006AFF}. Experimental results show that it can achieve state-of-the-art recognition performance, especially in the case of clothing and carrying backpacks.
}

\begin{figure*}
	\centering
	\includegraphics[scale=0.35]{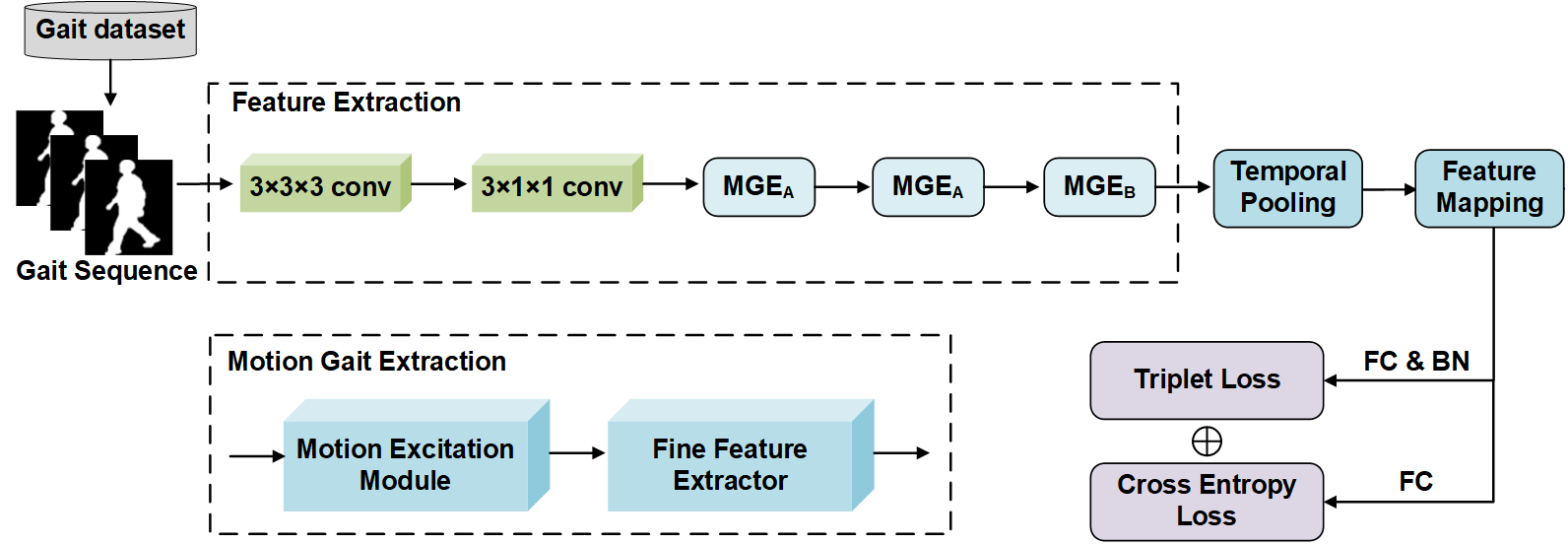}
	\caption{Overview of the proposed MotionGait, where FC represents separate fully connected layer and BN refers to batch normalization. $MGE_A$ and $MGE_B$ respectively correspond to different feature fusion modes of Fine Feature Extractor (FFE) module.}
	\label{FIG1}
\end{figure*}

\section{Related work}
\par{
	In this section, we discuss the related work in the field of action recognition and gait recognition based on motion features and spatio-temporal features respectively. Among them, ACTION-Net \cite{Wang2021ACTIONNetME} and GaitGL \cite{Lin2021GaitRV} inspired us to propose the Motion Gait.
}
\subsection{Action Recognition}
\par{
	In the field of action recognition, \cite{Simonyan2014TwoStreamCN} proposed the two-stream convolutional network in the early stage, in which the temporal flow is used to capture the motion information and temporal behavior of the object. I3D \cite{Carreira2017QuoVA} calculated the optical flow in advance and form the two-stream architecture with RGB, which is used to better represent the motion mode. However, the models of these two-stream architectures have a large number of parameters. Then, STM \cite{Jiang2019STMSA} proposed Channel-wise SpatioTemporal Module (CSTM) and Channel-wise Motion Module (CSM) to model spatio-temporal and motion information respectively, in which CSM learns motion information through the relationship between adjacent frames. In order to get attention to the temporal properties of video, TEA \cite{Li2020TEATE} introduced multiple temporal aggregation (MTA) and motion excitation (ME), which obtained motion information by convolution and subtraction of adjacent frames. Based on the work of TEA, ACTION-Net \cite{Wang2021ACTIONNetME} proposed several parallel excitation modules, in which the Motion Excitation (ME) module realized the effect similar to attention according to the change information of adjacent frames. However, the above method adds convolution in the calculation of adjacent frame relationship, which will distort the motion change information to a certain extent, and lose the corresponding relationship between the motion change and the original feature in spatio-temporal position.
}
\subsection{Gait Recognition}
\par{
	At present, the most advanced works \cite{Huang20213DLC,Lin2021GaitRV,Fan2020GaitPartTP} focus on the temporal and spatial attributes of gait, and some works \cite{Li2020GaitRV,Zhang2019GaitRV} aim at extracting invariant gait features and decoupling gait features. All these efforts are devoted to improving recognition performance under different covariates (cross-view, carrying objects and clothing changes).
}
\par{
	In terms of spatial gait features, Gaitset \cite{Chao2019GaitSetRG} proposed horizontal segmentation and used horizontal pyramid mapping to obtain multi-scale spatial representations. Subsequent works \cite{Wang2019LearningVI,Zhang2020CrossViewGR} also followed the practice of horizontal segmentation to obtain more refined local features, GaitPart \cite{Fan2020GaitPartTP} introduced focal convolution layer (FConv) to learn the local features after horizontal segmentation. Gaitgl \cite{Lin2021GaitRV} proposed global and local feature extractor to integrate global and segmented local representations. In addition, GLN \cite{Hou2020GaitLN} obtained more abundant spatial information by fusing multi-scale feature maps.
}
\par{
	In the aspect of temporal information utilization, many works \cite{Chao2019GaitSetRG,Hou2020GaitLN,Lin2021GaitRV} aggregated feature sequences in the temporal dimension (through temporal pooling or convolution) and fuse multi-temporal scales gait features in stages. Gaitpart \cite{Fan2020GaitPartTP} proposed micro-motion capture module to capture short-range temporal gait features. Zhang et al. \cite{Zhang2020CrossViewGR} applied LSTM attention module to the segmented parts to learn the temporal relationship.
}
\par{
	In addition, Lin et al. \cite{Lin2020GaitRW} used 3D convolution to conduct multi-scale spatio-temporal modeling of gait features, GaitGL \cite{Lin2021GaitRV} integrated multi-granularity spatial information, and used local temporal aggregation operation to preserve the integrity of spatial information. However, these methods rigidly learned motion patterns, ignored the motion changes during walking, and also lost the independence of different local parts to a certain extent. Therefore, we propose the novel Motion Excitation Module (MEM) to focus on the information of pedestrian movement changes, and improve the uniqueness of each local feature through the Fine Feature Extractor (FFE).
}
\begin{figure}
	\centering
	\includegraphics[scale=0.35]{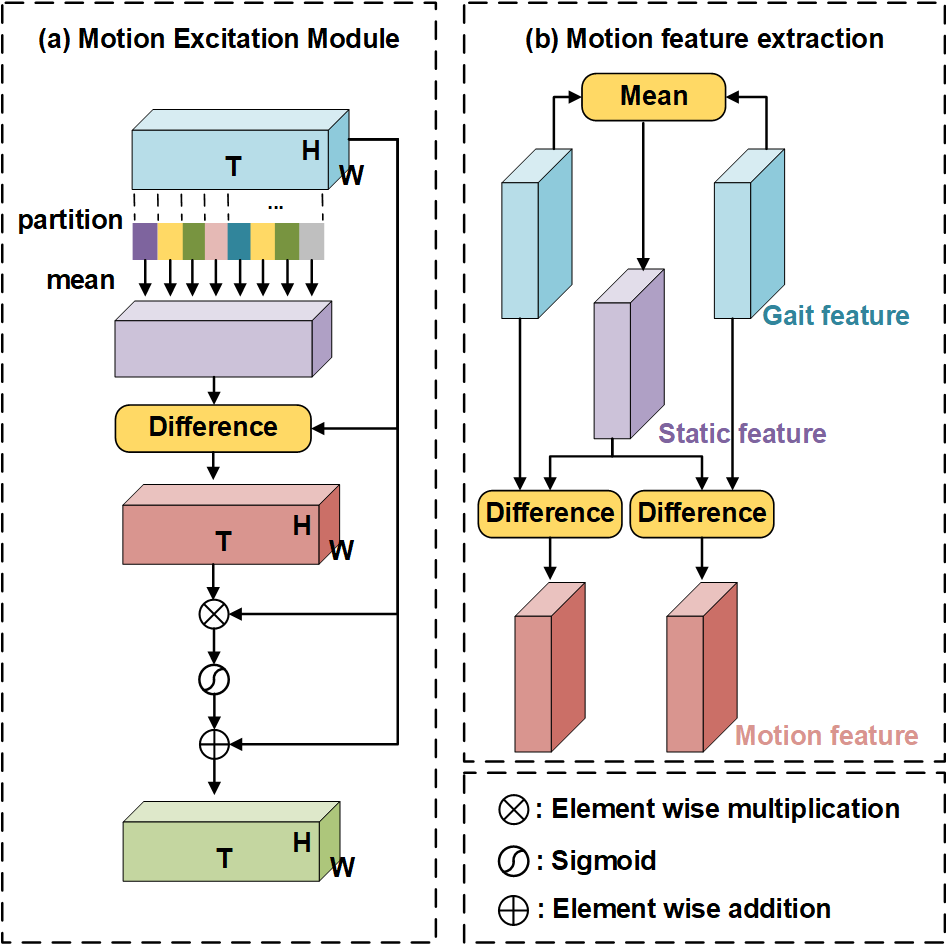}
	\caption{Illustration of Motion Excitation Module (MEM) and motion feature extraction.}
	\label{FIG2}
\end{figure}

\section{Proposed Method}
\par{
	In this section, we first introduce the framework of the proposed method. Then, we describe the key modules Motion Excitation Module (MEM) and Fine Feature Extractor (FFE). Finally, the details of training and testing are given.
}
\subsection{Overview}\label{section3.1}
\par{
	The overview of the proposed gait recognition method is shown in Fig.\ref{FIG1}(a). Firstly, convolution and LTA  \cite{Lin2021GaitRV} are used to extract shallow gait features from gait silhouette sequences. Next, spatio-temporal gait features with motion excitation are extracted through Motion Gait Extraction (MGE). MGE is mainly composed of Motion Excitation Module (MEM) and Fine Feature Extractor (FFE). According to the different fusion methods of FFE for global and local features, it can be divided into MGEA and MGEB, which represent element wise addition and concatenation of height dimension respectively. MEM mines the motion change information contained in the feature map itself as motion excitation. FFE retains the uniqueness of different local features and integrates global and local information. Then, temporal pooling and GeM pooling layer  \cite{Lin2021GaitRV} are used for feature mapping. Finally, our model is trained by cross entropy loss and triplet loss \cite{Chao2019GaitSetRG,Fan2020GaitPartTP}.
}

\subsection{Motion Excitation Module}\label{section3.2}
\par{
	Previous works in the field of motion recognition \cite{Jiang2019STMSA,Li2020TEATE,Wang2021ACTIONNetME} have divided the features in the temporal dimension, learned the changing motion information by doing difference after convolution, and usually extracted the motion change features through spatial pooling and convolution. However, the convolution operation will distort the motion change information and lose the corresponding relationship between the motion change feature and the original feature in time and space, which is particularly detrimental to the tasks with diverse and different motion changes. Considering that pedestrian gait has different amplitude and periodic motion changes in different local parts, we propose the Motion Excitation Module (MEM) to capture the temporal changes of gait features, which can preserve the original and diverse motion changes of features.
}
\par{
	The proposed MEM process is shown in Fig.\ref{FIG2}(a). Suppose the input is ${{X}_{in}}=\{{X}_{in}^{i} \vert i=1,..,s \} \in {R}^{{c}\times{s}\times{h}\times{w}}$, where c is the number of channels, s is the length of the sequence of the feature maps, and (h,w) represents the height and width of each frame. In the temporal dimension, we first divide the sequence ${X}_{in}$ into clips and $ {X}_{clip}=\{{{X}_{clip}}^{j} \vert j=1,...,(\lfloor {s/L} \rfloor+1)\}$, where L is a super parameter indicating the length of the clip sequence and $\lfloor \cdot \rfloor$ means integer down, ${X}_{clip}^j \in {R}^{{c}\times{L}\times{h}\times{w}}$. Then, we use statistical function to calculate the mean value of each clip as the static feature map, The formula can be expressed as follows
}
\begin{equation}
	\begin{aligned}
		{X}_{sta}^{j}={F_m}({X}_{clip}^{j})
	\end{aligned}
\end{equation}
\noindent where $F_m$ represents taking the average value on dimension s, ${X}_{sta}^{j} \in {R}^{{c}\times{1}\times{h}\times{w}}$.
\par{
	Then, the primary motion feature sequence is obtained by the difference between the static feature sequence and the original sequence, Fig.\ref{FIG2}(b). shows the processing flow of action features while L=2. Furthermore, considering the influence of chronological differences of different frames on the positive and negative values of elements, we calculate the absolute value as the final motion feature sequence after the difference operation. The above process can be expressed as follows
}
\begin{equation}
	\begin{aligned}
		{X}_{motion}^{i}=\vert {X}_{in}^{i}-{X}_{sta}^{\lceil i/L \rceil} \vert
	\end{aligned}
\end{equation}
\noindent where $\lceil \cdot \rceil$ represents rounding up to an integer, ${X}_{motion}^{i} \in {R}^{{c}\times{s}\times{h}\times{w}}$.
\par{
	Finally, the motion feature sequence ${{X}_{motion}}=\{{X}_{motion}^{i} \vert i=1,..,s \}$ and the original feature sequence ${X}_{in}$ are multiplied based on element, nonlinear mapping is added to improve the expression ability of features, and the residual connection is used to fuse with the original features. Gait features excited by motion features can be expressed as
}
\begin{equation}
	\begin{aligned}
		{X}_{MEM}={X}_{in}+{\delta}(M({X}_{in},{X}_{motion}))
	\end{aligned}
\end{equation}
\noindent where $\delta(\cdot)$ represents nonlinear activation function sigmoid, M is element based multiplication and ${{X}_{MEM}}=\{{X}_{MEM}^{i} \vert i=1,..,s \} \in {R}^{{c}\times{s}\times{h}\times{w}}$.
\par{
	Based on the above operations, MEM does not add additional parameters and can handle non fixed length sequences. Experimentally, MEM is placed before each Fine Feature Extractor (FFE) and together form Motion Gait Extractor.
}

\begin{figure*}[t]
	\centering
	\includegraphics[scale=0.5]{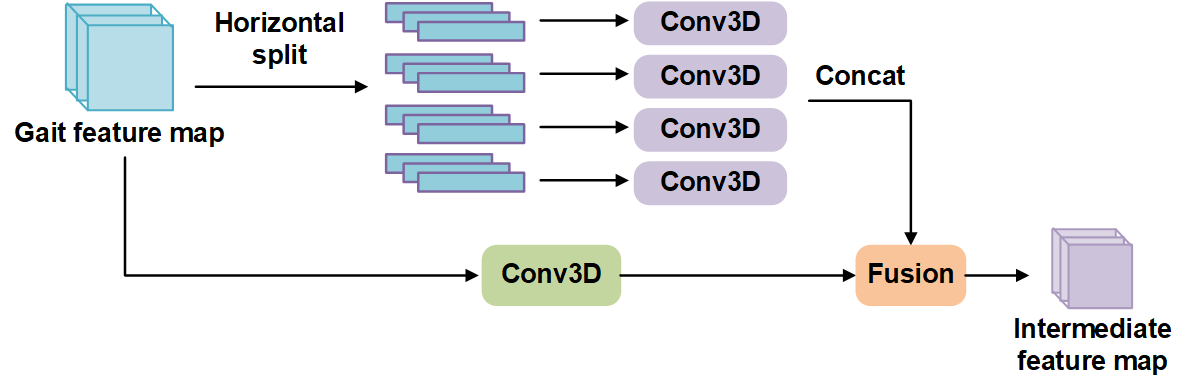}
	\caption{Structure of Fine Feature Extractor (FFE). The number of horizontal segmentation parts is controlled by the super parameter n, and the figure shows the case while n=4.}
	\label{FIG3}
\end{figure*}

\subsection{Fine Feature Extractor}\label{section3.3}
\par{
	In the process of pedestrian walking, different parts have various appearance and motion patterns, previous works \cite{Fan2020GaitPartTP,Chao2019GaitSetRG,Li2018DiversityRS,Luo2019BagOT,Sun2018BeyondPM,Li2018HarmoniousAN,Yao2019DeepRL,Su2017PoseDrivenDC} have also shown that feature representations of different parts can provide fine-grained information, so we introduce Fine Feature Extractor (FFE) to extract fine local features and retain the independence of different parts of information, the architecture of FFE is shown in Fig.\ref{FIG3}. Assuming that ${X}_{MEM} \in {R}^{{c}\times{s}\times{h}\times{w}}$ is the output of MEM, where c is the number of channels, s is the length of the sequence of the feature maps, and (H,W) is the space size of each frame. First, input features are processed by 3D convolution to obtain global gait features ${X}_{global}$, which are formulated as follows
}
\begin{equation}
	\begin{aligned}
		{X}_{global}={F}_{{3}\times{3}\times{3}}({X}_{MEM})
	\end{aligned}
\end{equation}
\noindent where ${F}_{{3}\times{3}\times{3}}(\cdot)$ represents 3D convolution with convolution kernel size of 3, and ${X}_{global} \in {R}^{{{c}^{\prime}}\times{s}\times{h}\times{w}}$.
\par{
	Secondly, for the branch of local feature extraction, we divide the input feature map ${X}_{MEM}$ horizontally into multiple local parts, then use multiple independent 3D convolutions to process features of these local parts, and finally concatenate the local features to obtain the fine feature ${X}_{local} \in {R}^{{{c}^{\prime}}\times{s}\times{h}\times{w}}$. The number of local parts is the same as the number of convolutions, which can ensure that the independence of different local information will not be interfered with each other. The process of extracting fine local features can be realized by the following formula
}
\begin{equation}
	\begin{aligned}
		{X}_{local}=Cat({f}_{{3}\times{3}\times{3}}^{k}({X}_{MEM}^{k}))
	\end{aligned}
\end{equation}
\noindent where $Cat(\cdot)$ represents horizontal splicing operation, ${f}_{{3}\times{3}\times{3}}^{k}(\cdot)$ represents the independent convolution of size 3, k marks the local part and the corresponding convolution, the value range of k is (1,n) and n represents the number of parts divided horizontally.
\par{
	Finally, FFE has two different ways of global and local feature fusion: element based addition and h-dimension splicing. Motion Gait Extraction (MGE) is also divided into $MGE_A$ and $MGE_B$ structures corresponding to the fusion methods of FFE respectively. Thus, the gait features after fusion are respectively expressed as follows
}
\begin{equation}
	\begin{aligned}
		{X}_{FFEA}={X}_{global}+{X}_{local}
	\end{aligned}
\end{equation}
\begin{equation}
	\begin{aligned}
		{X}_{FFEB}=Cat({X}_{global},{X}_{local})
	\end{aligned}
\end{equation}
\noindent where $Cat(\cdot)$ represents horizontal splicing operation, ${X}_{FFEA} \in {R}^{{{c}^{\prime}}\times{s}\times{h}\times{w}}$ and ${X}_{FFEB} \in {R}^{{{c}^{\prime}}\times{s}\times{2h}\times{w}}$.
\par{
	Experimentally, $MGE_B$ is used to implement the last MGE module, and the rest is realized by $MGE_A$.
}

\begin{table*}[]
	\caption{Averaged rank-1 accuracy (\%) on CASIA-B, excluding identical-view cases.}\label{tab1}
	\resizebox{\textwidth}{!}{%
		\begin{tabular}{cl|llllllllllll}
			\toprule
			\multicolumn{2}{l|}{Gallery NM\#1-4}                             & \multicolumn{12}{c}{0° - 180°}                                                                                                                                                                                                                                                                                                       \\ \hline
			\multicolumn{2}{c|}{Probe}                                       & \multicolumn{1}{c}{0°} & \multicolumn{1}{c}{18°} & \multicolumn{1}{c}{36°} & \multicolumn{1}{c}{54°} & \multicolumn{1}{c}{72°} & \multicolumn{1}{c}{90°} & \multicolumn{1}{c}{108°} & \multicolumn{1}{c}{126°} & \multicolumn{1}{c}{144°} & \multicolumn{1}{c}{162°} & \multicolumn{1}{c|}{180°}          & \multicolumn{1}{c}{Mean} \\ \hline
			\multicolumn{1}{c|}{\multirow{4}{*}{NM\#5-6}} & GaitSet\cite{Chao2019GaitSetRG}          & 90.8                   & 97.9                    & \textbf{99.4}           & 96.9                    & 93.6                    & 91.7                    & 95.0                     & 97.8                     & 98.9                     & 96.8                     & \multicolumn{1}{l|}{85.8}          & 95.0                     \\
			\multicolumn{1}{c|}{}                         & GaitPart\cite{Fan2020GaitPartTP}         & 94.1                   & \textbf{98.6}           & 99.3                    & 98.5                    & 94.0                    & 92.3                    & 95.9                     & 98.4                     & 99.2                     & 97.8                     & \multicolumn{1}{l|}{90.4}          & 96.2                     \\
			\multicolumn{1}{c|}{}                         & GaitGL\cite{Lin2021GaitRV}           & 96.0                   & 98.3                    & 99.0                    & 97.9                    & \textbf{96.9}           & 95.4                    & 97.0                     & \textbf{98.9}            & \textbf{99.3}            & \textbf{98.8}            & \multicolumn{1}{l|}{94.0}          & 97.4                     \\
			\multicolumn{1}{c|}{}                         & MotionGait(ours) & \textbf{96.1}          & 98.1                    & 99.0                    & \textbf{98.1}           & 96.8                    & \textbf{95.7}           & \textbf{97.5}            & \textbf{98.9}            & \textbf{99.3}            & 98.6                     & \multicolumn{1}{l|}{\textbf{94.6}} & \textbf{97.5}            \\ \hline
			\multicolumn{1}{c|}{\multirow{4}{*}{BG\#1-2}} & GaitSet\cite{Chao2019GaitSetRG}          & 83.8                   & 91.2                    & 91.8                    & 88.8                    & 83.3                    & 81.0                    & 84.1                     & 90.0                     & 92.2                     & 94.4                     & \multicolumn{1}{l|}{79.0}          & 87.2                     \\
			\multicolumn{1}{c|}{}                         & GaitPart\cite{Fan2020GaitPartTP}         & 89.1                   & 94.8                    & 96.7                    & 95.1                    & 88.3                    & \textbf{94.9}           & 89.0                     & 93.5                     & 96.1                     & 93.8                     & \multicolumn{1}{l|}{85.8}          & 91.5                     \\
			\multicolumn{1}{c|}{}                         & GaitGL\cite{Lin2021GaitRV}           & 92.6                   & \textbf{96.6}           & 96.8                    & 95.5                    & 93.5                    & 89.3                    & 92.2                     & 96.5                     & 98.2                     & \textbf{96.9}            & \multicolumn{1}{l|}{91.5}          & 94.5                     \\
			\multicolumn{1}{c|}{}                         & MotionGait(ours) & \textbf{93.8}          & 96.1                    & \textbf{97.1}           & \textbf{96.0}           & \textbf{95.0}           & 90.6                    & \textbf{94.4}            & \textbf{97.8}            & \textbf{98.3}            & 96.8                     & \multicolumn{1}{l|}{\textbf{92.7}} & \textbf{95.3}            \\ \hline
			\multicolumn{1}{c|}{\multirow{4}{*}{CL\#1-2}} & GaitSet\cite{Chao2019GaitSetRG}          & 61.4                   & 75.4                    & 80.7                    & 77.3                    & 72.1                    & 70.1                    & 71.5                     & 73.5                     & 73.5                     & 68.4                     & \multicolumn{1}{l|}{50.0}          & 70.4                     \\
			\multicolumn{1}{c|}{}                         & GaitPart\cite{Fan2020GaitPartTP}         & 70.7                   & 85.5                    & 86.9                    & 83.3                    & 77.1                    & 72.5                    & 76.9                     & 82.2                     & 83.8                     & 80.2                     & \multicolumn{1}{l|}{66.5}          & 78.7                     \\
			\multicolumn{1}{c|}{}                         & GaitGL\cite{Lin2021GaitRV}           & 76.6                   & 90.0                    & 90.3                    & 87.1                    & 84.5                    & 79.0                    & 84.1                     & 87.0                     & 87.3                     & 84.4                     & \multicolumn{1}{l|}{69.5}          & 83.6                     \\
			\multicolumn{1}{c|}{}                         & MotionGait(ours) & \textbf{77.7}          & \textbf{92.6}           & \textbf{94.1}           & \textbf{91.0}           & \textbf{86.1}           & \textbf{79.9}           & \textbf{85.4}            & \textbf{89.9}            & \textbf{92.2}            & \textbf{87.6}            & \multicolumn{1}{l|}{\textbf{73.5}} & \textbf{86.4}            \\ \bottomrule
		\end{tabular}%
	}
\end{table*}

\subsection{Implementation Details}\label{section3.4}
\par\noindent{
	{\bf Hyper-parameters. \rm} As shown in Fig.\ref{FIG2}(a), the input features are first divided into clips of the same length in MEM, and the clip are used as the basic unit to participate in the subsequent calculation of static features and motion features. The length L of the clip determines the temporal span of the captured motion change, the larger the L, the motion feature contains the change information with a larger temporal span. In order to extract accurate and short-term motion changes, we build the clip with two frames, i.e. L=2. In addition, according to the experimental proof of previous works \cite{Lin2021GaitRV,Chao2019GaitSetRG}, we set n to 8, which is the number of local parts divided horizontally in FFE.
}

\par\noindent{
	{\bf Loss and Sampler. \rm} In terms of loss function, we use Euclidean distance as the basis of feature similarity measurement, and use the sum of triple loss \cite{Hermans2017InDO} and cross entropy loss as the joint loss function. The cross entropy loss is used to expand the distinguishability between different samples, and the triple loss can effectively expand the distance between positive and negative sample pairs. Moreover, the margin for triple loss is set to 0.2. For sampling, we use the Batch ALL (BA) sampling method \cite{Chao2019GaitSetRG,Bashir2009GaitRU,Hermans2017InDO}, each batch includes P subjects, and each subject contains K samples, therefore the batch size is $P \times K$. In the training phase, the number of input frames is fixed to reduce memory consumption.
}
\par\noindent{
	{\bf Testing. \rm} In the test phase, the training set is divided into gallery set and probe set. The gallery set contains the feature representations from the standard view, and the probe set includes the feature representations of the samples to be identified. Euclidean distance is selected to calculate the similarity between gallery and test set.
}

\section{Experiments}
\subsection{Dataset}
\par\noindent{
	{\bf CASIA-B. \rm} CASIA-B dataset \cite{Yu2006AFF} is a widely used dataset with 11 different views and 124 subjects. Each subject in CASIA-B contains 10 groups of sequences with three different walking conditions, of which 6 groups are normal walking (NM), 2 groups are walking with backpacks (BG), and the other 2 groups are wearing coats or jackets (CL). Each group contains gait collected from different views. Specifically, there are 11 angles ranging from 0° to 180° with an interval of 18°. Therefore, CASIA-B has a total of 13640 gait sequences, which are divided into training set and test set. Following the popular protocol proposed by \cite{Fan2020GaitPartTP}, this paper takes the first 74 subjects as the training set and the other 50 subjects as the test set. During the test stage, 4 sequences (NM\#1-4) under NM condition were used as gallery set, and the remaining 6 sequences (NM\#5-6, BG\#1-2 and CL\#1-2) were used as probe set to evaluate the performance of the proposed model.
}

\subsection{Training Details}
\par{
	In the training stage, we use the proposed model to extract gait features from gait silhouette sequences, and then use the joint loss function to calculate the loss of gait feature representations. {\bf 1) \rm} Common configurations. We use the method mentioned in \cite{Chao2019GaitSetRG} to extract gait silhouettes from CASIA-B \cite{Yu2006AFF} dataset, and normalize the size of each input frame to $64 \times 44$. All experiments adopt the Adam optimizer with the learning rate of 1e-4. {\bf 2) \rm} In CASIA-B, we set both P and K of batch size to 8. The length of the input sequence is fixed at 30 in the training phase, and the complete gait silhouette sequence is used as the input of the proposed model in the test phase. In addition, epoch number is set to 90K.
}

\vspace{-0.8cm}
\begin{table}[]\centering
	\caption{Accuracy (\%) of different combinations of modules on CASIA-B. The result is the average accuracy of 11 views, excluding identical-view cases.}\label{tab2}
	\begin{tabular}{c|c|ccc}
		\toprule
		MEM & FFE        & NM            & BG            & CL            \\ \hline
		\checkmark   &            & 97.4          & 94.7          & 85.2          \\
		& \checkmark & 97.4          & 95.3          & 86.0            \\
		\checkmark   & \checkmark & \textbf{97.5} & \textbf{95.3} & \textbf{86.4} \\ \bottomrule
	\end{tabular}%
\end{table}
\vspace{-0.8cm}

\subsection{Comparison with State-of-the-Art Methods}
\par{
	{\bf CASIA-B. \rm}As shown in Tab.\ref{tab1}, we compare the proposed MotionGait with state-of-the-art gait recognition methods GaitSet \cite{Chao2019GaitSetRG}, GaitPart \cite{Fan2020GaitPartTP} and GaitGL \cite{Lin2021GaitRV} on CASIA-B. In the comparative experiment, the size of the input gait silhouette map is $64 \times 44$, and all cross-view and various walking conditions (NM, BG, and CL) are taken into account. It can be seen that, in the great majority views, our proposed method achieves the best recognition performance. According to different walking conditions (NM, BG, and CL), we further analyze the comparison results. Under cross-walking-condition cases, compared with the advanced GaitGL \cite{Lin2021GaitRV}, the average recognition accuracy of our method is improved by 0.8\% and 2.8\% respectively under BG and CL conditions. We believe that backpacks or coats will mask the movement pattern of pedestrians under cross-walking-condition cases, and the motion change information captured by our proposed method is less affected by appearance, so it can capture more robust gait features. Compared with BG and CL, our method has little improvement in NM condition. This may be because the spatio-temporal characteristics can well represent the pedestrian motion law when there are no factors interfering with the appearance change, and the improvement brought by the motion change information is limited. In terms of average accuracy, due to the full consideration of local features and the innovative use of motion change information, our method achieves the highest accuracy under all three conditions.
}

\vspace{-0.8cm}
\begin{table}[]\centering
	\caption{Ablation study on clip length in MEM on CASIA-B. n represents the super parameter of clip length.}\label{tab3}
	\resizebox{30mm}{!}{%
	\begin{tabular}{c|ccc}
		\toprule
		n & NM            & BG            & CL            \\ \hline
		2 & 97.5          & \textbf{95.3} & 86.4          \\
		3 & \textbf{97.6} & \textbf{95.3} & 86.3          \\
		4 & \textbf{97.6} & 95.0          & \textbf{86.5} \\
		5 & 97.6          & 95.0          & 86.5          \\ \bottomrule
	\end{tabular}%
}
\end{table}
\vspace{-0.8cm}

\subsection{Ablation Study}
\par{
	In order to verify the effectiveness of each module in MotionGait, we conducte ablation experiments with different settings on CASIA-B, including the addition of MEM and FFE separately and the change of clip length in MEM.
}

\par{
	{\bf Analysis of MEM. \rm}Different from the traditional gait recognition methods, our Motion Excitation Module (MEM) focuses on the motion change information of the pedestrian walking process, and combines the motion change features with the spatio-temporal features to obtain a more appropriate gait representation of the walking process. As shown in Tab.\ref{tab2}, we performed ablation experiment on the effectiveness of the proposed modules. It can be seen from the data that after adding MEM, the recognition accuracy of the model has been improved under all conditions (NM, BG and CL), especially 94.7\% and 85.2\% under BG and Cl respectively. These data prove that the motion change information extracted by MEM is robust and can reduce the impact of appearance changes such as clothing and carrying objects, so as to have better performance under cross-walking-condition cases.
}
\par{
	MEM is dedicated to extracting motion change information. The temporal span of the extracted information is affected by the super parameter n, which defines the length of the divided clip in the temporal dimension. As shown in Tab.\ref{tab3}, we conducted ablation experiment on the setting of the super parameter n. We can observe that n has an improvement effect under various values, and has achieved good performance in some cases (for example, while n=5, it reaches 86.5\% under CL condition), but while n=2, the model is relatively stable under all conditions. At the same time, in order to extract more accurate short-term motion changes and maintain the stability of the model, we finally chose n=2 to achieve MEM.
}
\par{
	{\bf Analysis of FFE. \rm}Fine Feature Extractor (FFE) learns according to different local parts, and convolutions of non-shared parameters can effectively learn different local motion patterns and appearance forms. As shown in Tab.\ref{tab2}, after adding FFE, the recognition accuracy in BG and CL cases reached 95.3\% and 86.0\% respectively, which further proves the effectiveness of learning different parts respectively. In the cross-walking-condition cases, compared with the hands, legs and feet, the trunk of the human body is greatly disturbed, and other unexpected conditions will also change the movement of some parts of the human body. Therefore, FFE can improve the identifiability and robustness of gait features by learning local parts separately.
}
\section{Conclusion}
\par{
	In this paper, we propose the gait recognition method MotionGait based on the fusion of motion change information and spatio-temporal features, which can generate the more robust gait representations. We introduce Motion Excitation Module to generate motion change information and stimulate spatiotemporal features to focus on different parts according to the degree of motion change. In addition, our Fine Feature Extractor can effectively focus on different local features and improve the independent expression of different parts. The core goal of our proposed method is to extract and utilize motion information and improve the robustness of gait features. Experiments on the popular public dataset CASIA-B \cite{Yu2006AFF} show the effectiveness of our proposed method.
}
\section*{Acknowledgement}
This work is supported by the Innovation Research  Funds for Shijiazhuang Tiedao University (No. YC2022057), the National Nature Science Foundation of China (No. 61972267), and the Nature Science Foundation of Hebei Province (No. F2019210306).
\clearpage
%
%
\bibliographystyle{splncs04}
\bibliography{arxiv}
\end{document}